\documentclass[iicol,sn-mathphys-num]{sn-jnl}% Math and Physical Sciences Numbered Reference Style 
\usepackage{graphicx}%
\usepackage{multirow}%
\usepackage{amsmath,amssymb,amsfonts}%
\usepackage{amsthm}%
\usepackage{mathrsfs}%
\usepackage[title]{appendix}%
\usepackage{xcolor}%
\usepackage{textcomp}%
\usepackage{manyfoot}%
\usepackage{booktabs}%
\usepackage{algorithm}%
\usepackage{algorithmicx}%
\usepackage{algpseudocode}%
\usepackage{listings}%
\theoremstyle{thmstyleone}%
\theoremstyle{thmstyletwo}%
\theoremstyle{thmstylethree}%

\usepackage{lipsum} 
\newcommand{\myparagraph}[1]{\vspace{8pt}\noindent\textbf{#1.}\ }

\begin{document}

%\title[Article Title]{Memory Based Foreground and Background Adaptation for Cross-Domain Object Detection}
\title[Article Title]{Visually Similar Pair Alignment for Robust Cross-Domain Object Detection}
%\title[Article Title]{Enhancing Cross-Domain Object Detection Through Alignment of Visually Similar Instances}


\author{\fnm{Onkar} \sur{Krishna}}\email{onkar.krishna.vb@hitachi.com}

\author{\fnm{Ohashi} \sur{Hiroki}}\email{hiroki.ohashi.uo@hitachi.com}

\affil{\orgdiv{Intelligent Vision Research Department}, \orgname{Hitachi Ltd.}, \orgaddress{\city{Kokubunji}, \postcode{185-8601}, \state{Tokyo}, \country{Japan}}}


\abstract{Domain gaps between training data (source) and real-world environments (target) often degrade the performance of object detection models.
Most existing methods aim to bridge this gap by aligning features across source and target domains but often fail to account for visual differences, such as color or orientation, in alignment pairs.
This limitation leads to less effective domain adaptation, as the model struggles to manage both domain-specific shifts (e.g., fog) and visual variations simultaneously. 
In this work, we demonstrate for the first time, using a custom-built dataset, that aligning visually similar pairs significantly improves domain adaptation.
Based on this insight, we propose a novel memory-based system to enhance domain alignment.
This system stores precomputed features of foreground objects and background areas from the source domain, which are periodically updated during training. 
By retrieving visually similar source features for alignment with target foreground and background features, the model effectively addresses domain-specific differences while reducing the impact of visual variations.
%the model focuses on domain-specific differences while minimizing distractions from visual variations. 
%
Extensive experiments across diverse domain shift scenarios validate our method's effectiveness, achieving 53.1\% mAP on Foggy Cityscapes and 62.3\% on Sim10k, surpassing prior state-of-the-art methods by 1.2\% and 4.1\% mAP, respectively.
}


\maketitle

\section{Introduction}\label{sec1}

Object detection models ~\cite{he2017mask, girshick2014rich, ren2015faster} have demonstrated strong performance on standard benchmark datasets~\cite{deng2009imagenet, lin2014microsoft, everingham2010pascal}. However, their ability to generalize to real-world environments remains limited. This is because these models often fail to adapt to new environments without being retrained on new data, a process that is both costly and time-consuming due to the necessity of manual labelling. This gap between training and real-world performance is a significant hurdle for deploying object detection systems in dynamic, real-world settings.

To address this challenge, Unsupervised Domain Adaptation (UDA)~\cite{carlucci2017autodial, lu2017unsupervised, tzeng2017adversarial} has emerged as a promising solution. UDA methods aim to reduce the domain gap by aligning features between a labeled source domain and an unlabeled target domain through adversarial learning, enabling models to adapt without requiring additional annotations. In cross-domain object detection this feature alignment occurs at both the image and instance levels, with instance-level alignment focusing on features extracted from object proposals generated by the detector.

In traditional approaches~\cite{he2019multi, rezaeianaran2021seeking}, instance-level alignment happens without considering object categories, so instances from different categories may be incorrectly aligned. For example, a cat from the target domain could be aligned with a person from the source domain. Such misalignment can result in poor knowledge transfer and suboptimal performance. Recent advancements, such as category-to-category (C2C) alignment methods~\cite{tian2021knowledge, xu2020cross, zhang2021rpn, zheng2020cross}, address this issue by ensuring that only instances from the same category are aligned.

Although C2C methods outperform traditional instance alignment techniques, they still have limitations. We argue that aligning a target instance to any arbitrary source instance within the same category is still suboptimal, as objects within the same category can differ significantly in visual appearance—such as variations in color and orientation. These visual differences complicate the domain adaptation process, forcing the model to handle both visual variations and domain-specific differences simultaneously. This dual burden detracts from the model's primary objective, which is domain alignment.

To address this problem, we proposed a method called MILA~\cite{krishna2023mila}, which incorporates a memory module to store precomputed source instance features. This memory, much larger than a mini-batch, increases the chances of finding visually similar source instances for alignment with target features. By selecting visually similar source-target pairs, even across different batches, MILA improves alignment by allowing the model to focus on domain-specific differences while minimizing the impact of irrelevant visual variations.

\begin{figure}[t!]
  \centering
  \includegraphics[width=\columnwidth]{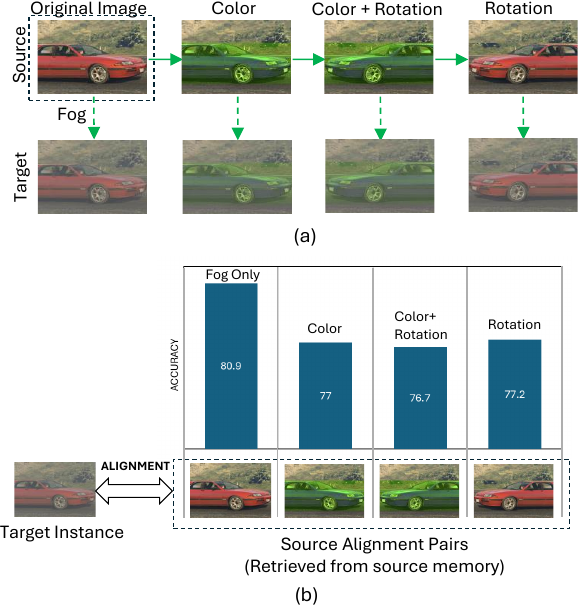} % fig111_ijcv.pdf
  \caption{(a) To validate our hypothesis, we introduce a new cross-domain dataset, AugSim10k → FoggyAugSim10k. The source dataset is created by augmenting Sim10k, applying new visual attributes such as color and orientation exclusively to the labeled objects. The target dataset is generated by applying fixed-intensity fog to the augmented source images. (b) Detection accuracy on the target dataset is compared using models trained with different instance alignment schemes. The results demonstrate that aligning visually similar pairs, differing only in domain characteristics (e.g., fog), significantly outperforms alignment of pairs with variations in color or orientation.}
  \label{fig:fig1}
\end{figure}

\myparagraph{Dataset Contribution}
This work builds on our previous work MILA, which achieved state-of-the-art (SOTA) results across five benchmark datasets. This success was driven by the key assumption that aligning visually similar instance pairs is crucial for effective domain alignment. However, despite its promising performance, this assumption had not been experimentally validated in our earlier work. 
In this paper, we address this gap by rigorously testing the hypothesis through the introduction of a novel cross-domain dataset. This dataset is specifically designed to control visual attributes such as color and orientation of labelled objects, allowing us to isolate and precisely measure the impact of visual similarity on alignment performance.
Our experiments, as shown in Fig.~\ref{fig:fig1}, demonstrate that aligning visually similar pairs leads to significantly improved performance compared to aligning pairs that differ in color, orientation, or both, confirming the pivotal role of visual similarity in domain adaptation. Furthermore, we make this customized dataset publicly available to facilitate future research in this area.

\myparagraph{Method Contribution} 
% We propose two key extension to MILA's network architecture to further enhance domain adaptation performance:
% (a) \textit{Background feature alignment}:
% MILA originally focused on aligning visually similar foreground instances but overlooked background features, which can carry important domain-specific information. For instance, in foggy environments, the background—such as the fog itself—plays a significant role in domain adaptation. To address this, we extend MILA by introducing a memory module that stores background features from source images and aligns them with visually similar backgrounds in target images. This dual alignment of both foreground and background features leads to a more comprehensive domain adaptation, especially when the background is important for distinguishing between domains.
% (b) \textit{SimSupConLoss}: We introduce SimSupConLoss, a new loss function that improves domain alignment by placing more emphasis on aligning visually similar pairs. The alignment is weighted by cosine similarity to ensure that similar pairs are aligned more effectively.
%
We propose a key extension to MILA's network architecture to further improve domain adaptation performance. While MILA originally focused on aligning visually similar foreground instances, it overlooked the role of background features (areas outside object bounding boxes), which often carry crucial domain-specific information. For example, in foggy environments, the background—such as the fog itself—can significantly influence domain adaptation. To address this, we extend MILA by introducing a memory module that stores background features from source images and aligns them with visually similar backgrounds in target images. This dual alignment of both foreground and background features leads to a more comprehensive domain adaptation, especially when the background is important for distinguishing between domains. Additionally, to manage memory efficiently and reduce redundancy during training, we introduce memory subsampling, ensuring optimal performance with minimal computational overhead.

\myparagraph{Experiment Contribution}
We evaluated the extended model and found that it further improves MILA’s performance, achieving new state-of-the-art results, such as a 4.1\% improvement on Sim10k and 2.5\% on Fogy Cityscapes datasets. 
\noindent We summarize our contributions as follows:
\begin{itemize}
     \item To the best of our knowledge, we are the first to demonstrate that aligning visually similar pairs enhances cross-domain object detection performance.

     \item To validate this hypothesis, we create and publicly release a novel cross-domain dataset with controlled visual attributes, such as object color and orientation.

     \item We extend MILA by enabling the alignment of both foreground and background features, leveraging visually similar backgrounds to enhance domain adaptation.

    % \item Generalization to unseen domains: We evaluate our method's ability to generalize effectively to previously unseen domains, showcasing its robustness across diverse environments.
\end{itemize}

%    \item We conduct extensive experiments on multiple benchmark scenarios, and the results demonstrate the effectiveness of our proposed method against MILA and previous state-of-the-art methods. Additionally, We perform new experiments to assess  ability of our proposed method to generalize to unseen domains, demonstrating its robustness in diverse environments.

% \noindent We summarize our contributions as follows:
% \begin{itemize}
%     \item \textbf{Experimental validation}: We are the first to show that aligning visually similar pairs improves cross-domain object detection performance.

%     \item \textbf{Dataset contribution}: We create and publicly release a novel cross-domain dataset with controlled visual attributes like color and orientation.

%     \item \textbf{Method extension}: We extend MILA to align both foreground and background features, improving domain adaptation by incorporating visually similar backgrounds.

%     \item \textbf{Generalization evaluation}: We assess MILA’s ability to generalize to unseen domains, demonstrating its robustness in diverse environments.
% \end{itemize}

\section{Related Works}\label{sec2}

\myparagraph{Object Detection} Object detection is the task of finding and labeling objects within an image. Current approaches can be broadly categorized into single-stage~\cite{redmon2016you, liu2016ssd} and two-stage models~\cite{ren2015faster}. While single-stage detectors are efficient and gaining popularity, two-stage detectors are still preferred for achieving higher performance. Faster R-CNN~\cite{ren2015faster} is a well-known two-stage detector and is favored for domain adaptive object detection due to its robustness and scalability. Following prior work~\cite{xu2020cross, he2020domain, zhang2021rpn, su2020adapting, zhu2019adapting}, we choose Faster R-CNN as our baseline in this study.
%\myparagraph{Object Detection.} Object detection involves identifying and classifying objects in an image. CNN-based detectors are typically categorized into single-stage~\cite{redmon2016you, liu2016ssd} and two-stage models~\cite{ren2015faster}. Two-stage models generally provide better detection performance and are more versatile for tasks like object segmentation~\cite{he2017mask, cheng2021dual}, thanks to the intermediate proposals they generate. In line with prior approaches~\cite{xu2020cross, he2020domain, zhang2021rpn, su2020adapting, zhu2019adapting}, we employ the widely-used two-stage network, Faster R-CNN~\cite{ren2015faster}, for its flexibility.

\myparagraph{Unsupervised Domain Adaptation (UDA)}
UDA is designed to address distribution shifts between different domains. It has been extensively studied across various computer vision tasks, such as image classification~\cite{wei2021metaalign, gao2021gradient}, semantic segmentation~\cite{hoyer2022daformer, chen2022deliberated}, and object detection~\cite{chen2018domain, saito2019strong, xu2020exploring}. Earlier UDA approaches aimed at reducing domain discrepancies in the feature space by optimizing specific metrics, including Maximum Mean Discrepancy (MMD)~\cite{gretton2012kernel, tzeng2014deep}, Weighted MMD~\cite{yan2017mind}, MultiKernel MMD~\cite{pei2018multi} and Wasserstein Distance~\cite{shen2018wasserstein}. More recently, domain adversarial learning has been introduced to further enhance UDA performance~\cite{tzeng2017adversarial, pei2018multi, saito2018maximum, wang2019transferable, chen2019progressive}. In this work, we focus specifically on domain adaptation for object detection.

%Adversarial Adaptation~\cite{tzeng2017adversarial, pei2018multi, saito2018maximum, wang2019transferable, chen2019progressive} introduces a domain discriminator to distinguish the source from the target, then feature extractor is encouraged to learn domain invariant features by fooling the discriminator. 

\myparagraph{Cross-domain Object Detection} 
Due to the localized nature of object detection, current methods often reduce domain disparity at multiple levels using adversarial feature adaptation, focusing on both image and instance alignment. DA-Faster~\cite{chen2018domain} was an early approach that aligned features at the image and instance levels. MAF~\cite{he2019multi} and~\cite{xie2019multi} expanded this idea by applying multi-layer feature adaptation across the backbone network. SWDA~\cite{saito2019strong} emphasized that strong local feature alignment is more effective than focusing on global alignment. CRDA~\cite{xu2020exploring} and MCAR~\cite{zhao2020adaptive} introduced multi-label classifiers to regulate features more effectively. Recent approaches~\cite{xu2020cross, he2020domain, li2020spatial, zhang2021rpn, vs2021mega, su2020adapting, zhu2019adapting} have focused on aligning instance-level features in a category-aware manner (C2C). These methods create a prototype for each category by aggregating multiple instances before alignment. However, collapsing all instances into a single prototype can cause a loss of intra-class variance, leading to sub-optimal alignment.

%It aims at improving performance of a detector, trained on the labeled source domain, on the new target domain. Due to its importance, numerous approaches have been proposed. One key trend is to align the feature distribution with adversarial learning. The main idea is to design an effective discriminator on mainly two feature spaces,  image-level [29, 26, 18, 17] and instance-level [5, 50, 12, 55, 52, 58] for detection. 

\myparagraph{Memory-based Cross-domain Detection}
Memory modules are commonly used in vision tasks, such as video object segmentation~\cite{rodriguez2019domain, oh2019video}, movie understanding~\cite{na2017read}, and visual tracking~\cite{yang2018learning}, for their ability to store and retrieve diverse types of knowledge. They have also been applied in domain adaptation~\cite{memsac} and cross-domain object detection~\cite{vs2021mega}.
MeGA-CDA~\cite{vs2021mega} is the closest work to ours, as it employs memory modules to store class prototypes and generate category-specific attention maps for enhanced category-to-category (C2C) alignment between source and target instances.
However, while both methods use memory, their objectives differ significantly. MeGA-CDA focuses solely on aligning paired instances of same category, whereas our method takes it further by considering the unique characteristics of individual instances, such as color and orientation, within each category. This finer-grained alignment makes our method more precise and effective for domain adaptation.

%Our method, on the other hand, aims to find the most visually similar source instance within a category for alignment. By aligning pairs that mainly differ in domain rather than visual characteristics, our approach allows the model to concentrate on the core task of domain alignment without being distracted by visual variations within the same category.  %which can be a limitation in MeGA-CDA.

%makes our approach more precise, not only aligns categories but also takes into account the finer details of individual instances within each category, leading to more effective domain alignment.
%\myparagraph{Differences from other works} 

\begin{figure*}[t!]
  \centering
  \includegraphics[scale=0.9, width=\textwidth]{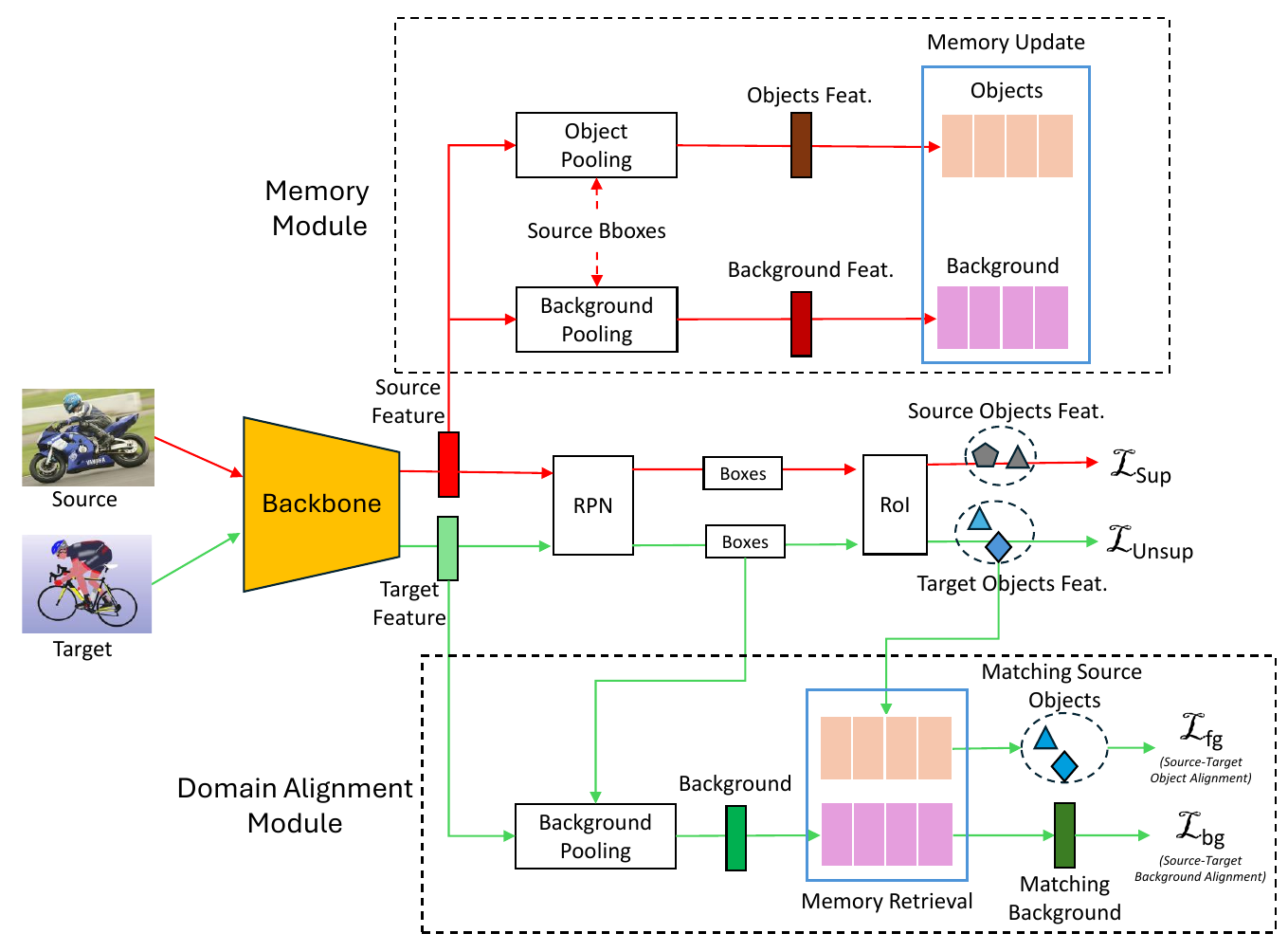} %network1.pdf% Adjust scale factor as needed
  \caption{Network Overview: Mainly consists of a \textbf{memory module}, a visual similarity-based foreground and background \textbf{domain alignment} module.}
  \label{fig:fig2}
\end{figure*}

\section{Method}\label{sec3}

\subsection{Preliminaries and Overview}\label{sec:overview}
We are given two datasets: a labeled source dataset \( \mathcal{D}_S = \left\{ \left( x_i^S, b_i^S, c_i^S \right) \right\}_{i=1}^{N_S} \), where \( x_i^S \) represents the source images, \( b_i^S \) the ground truth bounding boxes, and \( c_i^S \) the class labels. Each bounding box corresponds to one of \( C \) object categories. The second dataset is an unlabeled target dataset \( \mathcal{D}_T = \left\{ x_j^T \right\}_{j=1}^{N_T} \), where \( x_j^T \) represents the target images, with no bounding box and label annotations.
The goal is to train a domain-invariant object detector using the labeled source dataset \( \mathcal{D}_S \) and the unlabeled target dataset \( \mathcal{D}_T \). Although \( \mathcal{D}_S \) and \( \mathcal{D}_T \) share the same label space, they are drawn from distinct data distributions, presenting significant challenges for UDA.

In this work we start by verifying our key assumption: aligning visually similar pairs is crucial for effective domain alignment. To validate this, we introduce a new cross-domain dataset. After confirming the assumption, we design a network (see Fig.~\ref{fig:fig2}) around this idea with two main components: 1) \textit{a memory module} that stores features of labeled objects from the source images in foreground memory, while the rest of the features go into background memory. 2) \textit{a domain alignment module} that matches target features (foreground and background) with visually similar features extracted from the source memory. %, using our proposed \textbf{SimSupConLoss}. 
More details about these modules are provided in the next sections.

\subsection{Validating Assumption}

\myparagraph{Dataset Preparation} %To test our assumption that aligning visually similar pairs enhances domain alignment, we modify the Sim10k dataset, which contains 10,000 images and 58,701 car bounding boxes. .. This forms the cross-domain dataset: AugSim10k $\rightarrow$ FoggyAugSim10k.
%  This dataset is specifically designed to control visual attributes such as color and orientation of labelled objects, allowing us to isolate and precisely measure the impact of visual similarity on alignment perfor  --
%
To validate our assumption that aligning visually similar pairs enhances domain alignment, we prepare a cross-domain dataset: AugSim10k $\rightarrow$ FoggyAugSim10k. 
This dataset is prepared by controlling visual attributes of the labeled objects, such as color and orientation, to evaluate the impact of differences in these attributes between alignment pairs on domain adaptation.
We begin by modifying the Sim10k dataset, which contains 10,000 images and 58,701 car bounding boxes. Transformations are applied only to the labeled objects, leaving the background unchanged. As shown in Fig.~\ref{fig:fig1}(a), we generate three variations for each image: (1) Color Transformed, where the object's colors are changed but their positions remain the same, (2) Color + Rotation, where the color-transformed objects are horizontally flipped, and (3) Rotation Only, where the original objects are flipped without altering their color. 
These transformations expand Sim10k fourfold into AugSim10k, but we randomly augment only 2,500 images to maintain a consistent size of 10,000. Fixed-intensity fog is then applied to AugSim10k, creating the target dataset, FoggyAugSim10k.

%  Together, these form the cross-domain dataset: AugSim10k $\rightarrow$ FoggyAugSim10k.
%We then apply fixed-intensity fog to AugSim10k, creating the target dataset FoggyAugSim10k. 

\myparagraph{Experimental Setup} With the dataset ready, we evaluate MILA, a Faster R-CNN-based framework for cross-domain object detection. MILA works in two stages. First, it creates a source memory by extracting and storing pooled features of all labeled objects from the source dataset using a pretrained Faster R-CNN. In the second stage,  domain alignment is performed by matching target proposals with the most visually similar source proposals stored in memory.
As shown in Fig.~\ref{fig:fig1}(b), we test four alignment strategies with MILA on our dataset. %AugSim10k $\rightarrow$ FoggyAugSim10k. 
In \textbf{Domain-Only (Fog) Alignment}, the target object is aligned with its exact counterpart from the source, differing only in the fog. In \textbf{Color-Difference Alignment}, the target is aligned with a color-transformed version of the source object to evaluate the impact of color changes. In \textbf{Rotation-Only Alignment}, the target is paired with a rotated version of the source object, assessing the effect of orientation changes. Lastly, in \textbf{Color + Rotation Alignment}, the target is matched with a source object that is both color-transformed and rotated, assessing the combined impact of these changes.

\myparagraph{Results} The results, shown in Fig.~\ref{fig:fig1}(b), demonstrate that MILA achieves the highest object detection performance in Domain-Only Alignment mode, with an accuracy of 80.9\%, a 4.2\% improvement over Color + Rotation Alignment mode. This supports our hypothesis that aligning visually similar pairs leads to better domain alignment, thereby improving detection performance.

 %In comparison, the accuracy for \textbf{Color-Difference Alignment} was 77.0\%, for \textbf{Rotation-Difference Alignment} 77.2\%, and for \textbf{Color + Rotation Alignment} 76.7\%.

\subsection{Memory Module}\label{sec:memory_creation}
%Based on our core hypothesis we design our network architecture. Most of the existing methods rely on mini-batch source-target pairs for domain alignment, however, if we rely on mini-batch based method we cant even guarantee to get same category objects in both source-target and its even rare to get visually similar pairs of same category for alignment. To address this issue, we propose constructing two types of memory using the labeled source dataset \( \mathcal{D}_S \). (i) \textit{foreground memory } stores features associated with labeled bounding boxes in the source images, along with their corresponding class labels. (ii) \textit{background memory} captures features from regions outside the ground truth bounding boxes, representing the background information.
%
%We design our network based on the core hypothesis that aligning visually similar pairs improves domain adaptation. Current methods often rely on mini-batch source-target pairs, which rarely include same-category, visually similar instances needed for effective alignment. To address this limitation, we propose constructing two types of memory using the labeled source dataset

Our network is designed around the core hypothesis that aligning visually similar pairs enhances domain adaptation. To achieve this, we introduce two types of memory constructed using the labeled source dataset \( \mathcal{D}_S \). The \textbf{foreground memory}, which stores pooled features of labeled bounding boxes from source images along with their corresponding class labels. The \textbf{background memory} which captures features from regions outside the ground truth bounding boxes, representing background information. This memory-based approach enables the model to identify and align visually similar pairs across domains, facilitating more effective domain alignment.

\myparagraph{Foreground Memory} 
%Given the labeled source dataset \( \mathcal{D}_S \), we extract instance-level features from the bounding boxes in the source images.
For each source image \( x_i^S \) with \( K_i^S \) labeled objects with there bounding boxes \( b_i^S = \left\{ b_{i,k}^S \right\}_{k=1}^{K_i^S} \), we process the image as follows: 

\begin{enumerate}%[label=(\roman*), leftmargin=*]
    \item \textbf{Feature Extraction:} The image is passed through a pre-trained Faster R-CNN backbone (trained on the source dataset) to generate a feature map \( f(x_i^S; \theta) \).
    
    \item \textbf{Pooling Features:} Using the bounding boxes, features for each object are extracted from the feature map through a Box Pooler:
    \[
    \text{pooled\_feat}_{i,k}^S = \text{BoxPooler}(f(x_i^S; \theta), b_{i,k}^S)
    \]

    % IJCV fix: here we also filter these boxes  
    
    \item \textbf{Detection Head Processing:} The pooled features are passed through the detection head to obtain object-specific features:
    \[
    g_{i,k}^S = \text{det}(\text{pooled\_feat}_{i,k}^S; \psi)
    \]
    
    \item \textbf{Foreground Memory Construction:} These object-specific features \( g_{i,k}^S \), along with their class labels \( c_{i,k}^S \), are stored in the foreground memory \( \mathcal{M}_\text{fg} \):
\end{enumerate}
 \[
    \mathcal{M}_\text{fg} = \{ (g_{i,k}^S, c_{i,k}^S) \mid i = 1, \dots, N_S; k = 1, \dots, K_i^S \}
    \]

\myparagraph{Background Memory} 
To extract the background features from source images, we reuse the feature map \( f(x_i^S; \theta) \), which is computed during the foreground feature extraction. Then, for each source image, we generate a binary mask based on the labeled bounding boxes:
\[
\text{mask}(b_i^S) = \sum_{k=1}^{K_i^S} \text{mask}(b_{i,k}^S)
\]
%This mask isolates the background by excluding the areas covered by the bounding boxes.  
The background features are then extracted by applying the mask to the feature map:
%\[
%\text{masked\_feat}_i^S = f(x_i^S; \theta) \times \left( 1 - \text{mask}(b_i^S) \right)
%\]
\[
\text{masked\_feat}_i^S = f(x_i^S; \theta) \times \left( 1 - \text{mask}(b_i^S) \right)
\]
Since these feature maps vary in size based on the number and dimensions of bounding boxes, we use Adaptive Pooling to create a fixed-size output of \( (7,7) \), which is then fed into the detection head:
\[
bg_i^S = \text{det}(\text{AdaPool}(\text{masked\_feat}_i^S, (7,7)); \psi)
\]
%Finally, the background memory is defined as:
The resulting background memory is composed of these pooled background features from all source images:
\[
\mathcal{M}_\text{bg} = \left\{ bg_i^S \mid i = 1, \dots, N_S \right\}
\]

\subsection{Domain Alignment Process}\label{sec:domain_alignment}

After constructing the source memory, domain adaptation training for each target image proceeds as follows: First, bounding boxes are detected in the target image, and low-confidence predictions are filtered out. Next, features are extracted for the filtered foreground objects as well as the background. Then, visually similar features are retrieved from the source memory. Finally, the target features are aligned with the retrieved source features to facilitate effective domain adaptation.

%After the memory creation, the domain adaptation training for a target image begins with the following steps: 1) predicting and filtering target bounding boxes, 2) extracting foreground and background target features for the filtered boxes, 3) retrieving visually similar features from the source memory, and 4) aligning the foreground and background target features with the retrieved source features. 

%\myparagraph{Target Foreground and Background Features}
\myparagraph{Target Feature Extraction}
For each target image \( x_j^T \), the Faster R-CNN predicts bounding boxes and class labels, denoted as:
\[
\hat{\mathcal{B}}_T = \left\{ \left( \hat{b}_{j,k}^T, \hat{c}_{j,k}^T, s_{j,k}^T \right) \mid k = 1, \dots, K_j^T \right\}
\]
Here, \( \hat{b}_{j,k}^T \) is the predicted bounding box, \( \hat{c}_{j,k}^T \) is the predicted class, and \( s_{j,k}^T \) is the confidence score for the \( k \)-th prediction. 
%box in the \( j \)-th target image. 
To remove inaccurate predictions, we apply non-maximum suppression (NMS) and confidence thresholding:
\[
\hat{\mathcal{B}}_T' = \text{NMS}(\hat{\mathcal{B}}_T), \quad \mathcal{B}_T' = \left\{ \hat{b}_{j,k}^T \mid s_{j,k}^T \geq \delta \right\}
\]
Next, for the filtered bounding boxes \( \mathcal{B}_T' \), we extract foreground features by pooling from the Faster R-CNN backbone feature map \( f(x_j^T; \theta)\), followed by processing these pooled features through the detection head:
\[
g_{j,k}^T = \text{det}(\text{BoxPooler}(f(x_j^T; \theta), \hat{b}_{j,k}^T); \psi)
\]
The background feature is obtained by masking out regions of \( f(x_j^T; \theta) \) corresponding to the filtered boxes, pooling the remaining features, and processing them with the detection head:
\[
bg_{j}^T = \text{det}(\text{AdaPool}(\text{masked\_feat}_{j}^T, (7,7)); \psi)
\]
The extraction process for target features is consistent with the steps used for creating the source memory, ensuring consistency across both.

%Foreground and background feature extraction for target images follows the same approach as used for memory creation with source images, ensuring consistency across both.

%This process follows the same feature extraction method used during memory creation for the source images, maintaining consistency in extracting foreground and background features from both source and target images.

\myparagraph{Memory Reterival}
For each target foreground feature \( g_{j,k}^T \), we retrieve the most similar positive sample from the same category in the source memory by maximizing cosine similarity:
\begin{equation}
g_{j,k}^{S+} = \arg\max_{g_{i,k}^S} \frac{g_{j,k}^T \cdot g_{i,k}^S}{\| g_{j,k}^T \| \| g_{i,k}^S \|}
\end{equation}
Similarly, for each target background feature  \( bg_i^S \), we similarly retrieve the most similar positive sample from the source background memory:
\begin{equation}
bg_{j}^{S+} = \arg\max_{bg_i^S} \frac{bg_j^T \cdot bg_i^S}{\| bg_j^T \| \| bg_i^S \|}
\end{equation}
Negative samples for both foreground and background features \( (g_{j, k}^{S-}, bg_{j}^{S-}) \) are obtained by randomly selecting one sample from categories different from the category of the positive pairs.
%This step ensures that both foreground and background target features are aligned with their visually similar counterparts from the source memory.

%Once we retrieve positive and negative set of instances from source memory for a given target instance,

\myparagraph{Foreground Alignment}
After retrieving positive and negative samples \( (g_{j, k}^{S+}, g_{j, k}^{S-}) \) for a target feature  \( g_{j, k}^{T}\), we align them using a specially designed triplet loss  \( \mathcal{L}_{\text{fg}}\). The loss is defined as:
% \[
% \mathcal{L}_{\text{triplet}} = \frac{1}{N} \sum_{j,k} w_{j,k} \cdot \left[ \| g_{j,k}^T - g_{j,k}^{S+} \|_2^2 - \min\left(\| g_{j,k}^T - g_{j,k}^{T-} \|_2^2\right) + \alpha \right]_+
% \]  
\begin{align*}
\mathcal{L}_{\text{fg}} &= \frac{1}{N} \sum_{j,k} w_{j,k} \cdot \Big[ \| g_{j,k}^T - g_{j,k}^{S+} \|_2^2 \\
&\quad - \min\big(\| g_{j,k}^T - g_{j,k}^{T-} \|_2^2\big) + \alpha \Big]_+
\end{align*}
where \( [\cdot]_+ \) denotes the ReLU operation to ensure non-negativity, \( \alpha \) is the margin hyperparameter, and \( w_{j,k} = \text{cosine}(g_{j,k}^T, g_{j,k}^{S+}) \) is similarity between the target feature and the positive source feature, used as a weight.
This formulation ensures that visually similar pairs across domains are aligned more strongly, improving domain adaptation.

\myparagraph{Background Alignment}
The target background feature \( \mathbf{bg}_j^T \) is aligned with the retrieved visually similar source background features \( \mathbf{bg}_j^S \) through adversarial domain adaptation. A binary domain discriminator, \( d(\cdot; \theta_d) : Z \to \{0, 1\} \), is trained to map background features to their respective domain labels. Consequently, the feature extractor learns to fool the discriminator by making features from both domains as similar as possible.

The novelty of our approach lies in aligning background features based solely on domain differences, avoiding irrelevant variations. This focus reduces the risk of aligning insignificant background discrepancies, which could otherwise lead to suboptimal alignment.
The adversarial domain alignment loss, \( \mathcal{L}_{bg} \), is defined as:

\[
\mathcal{L}_{bg} = - \mathbb{E}_{\mathbf{bg}_j^S \sim \mathcal{D}_S} \left[ \log d(\mathbf{bg}_j^S; \theta_d) \right]
\]

\[
- \mathbb{E}_{\mathbf{bg}_j^T \sim \mathcal{D}_T} \left[ \log(1 - d(\mathbf{bg}_j^T; \theta_d)) \right]
\]

\noindent In this process, gradient reversal is applied to both the source and target features before passing them to the discriminator.

\subsection{Overall Objective}

In addition to the domain alignment losses \( \mathcal{L}_{fg} \) and \( \mathcal{L}_{bg} \) introduced in the previous section, we include supervised and unsupervised losses in our overall objective function. The supervised loss \( \mathcal{L}_{Sup} \) is the standard object detection loss optimized using the labeled source domain dataset, while the unsupervised loss \( \mathcal{L}_{Unsup} \) is calculated on target images with pseudo labels as described in~\cite{li2022cross}. 
Combining these components, the overall objective function is defined as:

\begin{equation}
\mathcal{L} = \mathcal{L}_{Sup} + \lambda_1 \mathcal{L}_{Unsup} + \lambda_2 \mathcal{L}_{fg} + \lambda_3 \mathcal{L}_{bg},
\label{eq:overall_loss}
\end{equation}
where \( \lambda_1, \lambda_2, \lambda_3 \) are hyperparameters that control the weight of each loss component.

\section{Experiments}
\subsection{Datasets}
We conducted extensive experiments on five public datasets across three domain shift scenarios, following the standard UDA setting in the literature~\cite{saito2019strong, li2022cross}.

\myparagraph{\textbf{Adverse Weather Adaptation}} 
In this scenario, we use the Cityscapes dataset~\cite{cordts2016cityscapes} as our source domain, consisting of 3,475 real urban images, with 2,975 for training and 500 for validation across eight object categories. For the target domain, we use Foggy Cityscapes~\cite{sakaridis2018semantic}, a synthetic variant of Cityscapes that simulates foggy conditions. Evaluation results are reported on the Foggy Cityscapes validation set.

\myparagraph{\textbf{Synthetic to Real Adaptation}} 
Sim10k~\cite{johnson2016driving} is a synthetic dataset generated from the game Grand Theft Auto V, containing 10,000 images with 58,701 annotated car bounding boxes. To adapt these synthetic scenes to real-world images, we use the entire Sim10k dataset as the source domain and the Cityscapes training set~\cite{cordts2016cityscapes} as the target domain. Since only the \textit{Car} class is annotated in both datasets, we evaluate our model’s performance on \textit{Car} detection using the Cityscapes validation set.

\myparagraph{\textbf{Real to Artistic Adaptation}} 
In this scenario, we evaluate our model's effectiveness in bridging the significant domain gap between real and artistic images. For the source domain, we use the Pascal VOC ~\cite{everingham2010pascal} dataset, which consists of 16,551 images across 20 common object categories. For the target domains, we utilize Comic2k~\cite{inoue2018cross} which includes 1,000 training and 1,000 test images in comic style, with 6 categories overlapping with those in Pascal VOC.

\subsection{Implementation Details}\label{sec:implementation_details}
%
% IJCV - add information about filtering memory 
Following prior works~\cite{saito2019strong, xu2020cross, tian2021knowledge, li2022cross}, we use Faster R-CNN~\cite{ren2015faster} as our base detection model, with either ResNet-101~\cite{he2016deep} or VGG16~\cite{simonyan2014very} (on Cityscapes) as the backbone
%, depending on the setting. The ResNet-101 backbone is fine-tuned from ImageNet~\cite{deng2009imagenet} pre-trained weights, while VGG16 is trained from scratch. 
As standard~\cite{he2019multi}, all images are resized to have a shorter side of 600 pixels while preserving aspect ratios. We apply both strong and weak data augmentations as described in~\cite{li2022cross}.
For evaluation, we report average precision (AP) for each class and the mean AP (mAP) across all classes. Unless specified otherwise, the hyperparameters are set as follows: $\lambda_1$=1.0, $\lambda_2$=0.05, and $\lambda_3$=0.05. The foreground and background memory are initialized once and updated at regular interval, with each memory slot storing features of dimension 1024. To filter noisy predictions on target images, we use a confidence threshold \(\delta\)=0.8 and set the margin \( \alpha \) to 1.5 for the  triplet loss  \( \mathcal{L}_{\text{fg}}\).
The model is trained using stochastic gradient descent (SGD) with a momentum of 0.9 and a fixed learning rate of 0.01, without learning rate decay. Our implementation builds on the code from~\cite{li2022cross}, following the same settings for other hyperparameters.
All experiments were run on 2 Nvidia V100 GPUs, with batch sizes of 4, using PyTorch and Detectron2.

%---------------------------------------------------------------
% CITYSCAPES
\begin{table*}[t]
 %\scriptsize
 \footnotesize	
  % \begin{center}
  \centering
  \addtolength{\tabcolsep}{-2.0pt}
  \scalebox{0.9}{
    % \resizebox{\textwidth}{!}{
    \begin{tabular*}{\textwidth}{@{\extracolsep{\fill}\quad} l|cccccccc|r}
      \toprule % <-- Toprule here
      Method & bus & bicycle & car & mcycle & person & rider & train & truck & mAP\\
      \midrule % <-- Midrule here
      Source (F-RCNN) & 20.1 & 31.9 & 39.6 & 16.9 & 29.0 & 37.2 & 5.2 & 8.1 & 23.5 \\
       \midrule 
      SCL~\cite{shen2019scl} & 41.8 & 36.2 & 44.8 & 33.6 & 31.6 & 44.0 & 40.7 & 30.4 & 37.9 \\
      DA-Faster~\cite{chen2018domain} & 35.3 & 27.1 & 40.5 & 20.0 & 25.0 & 31.0 & 20.2 & 22.1 & 27.6 \\
      SCDA~\cite{zhu2019adapting} & 39.0 & 33.6 & 48.5 & 28.0 & 33.5 & 38.0 & 23.3 & 26.5 & 33.8 \\
      SWDA~\cite{saito2019strong} & 36.2 & 35.3 & 43.5 & 30.0 & 29.9 & 42.3 & 32.6 & 24.5 & 34.3 \\
      DM~\cite{kim2019diversify} & 38.4 & 32.2 & 44.3 & 28.4 & 30.8 & 40.5 & 34.5 & 27.2 & 34.6 \\
      MTOR~\cite{cai2019exploring} & 38.6 & 35.6 & 44.0 & 28.3 & 30.6 & 41.4 & 40.6 & 21.9 & 35.1 \\
      MAF~\cite{he2019multi} & 39.9 &  33.9 & 43.9 & 29.2 & 28.2 & 39.5 & 33.3 & 23.8 & 34.0 \\
      iFAN~\cite{zhuang2020ifan} & 45.5 & 33.0 & 48.5 & 22.8 & 32.6 & 40.0 & 31.7 & 27.9 & 35.3 \\
      CRDA~\cite{xu2020exploring} & 45.1 & 34.6 & 49.2 & 30.3 & 32.9 & 43.8 & 36.4 & 27.2 & 37.4 \\
      HTCN~\cite{chen2020harmonizing} & 47.4 & 37.1 & 47.9 & 32.3 & 33.2 & 47.5 & 40.9 & 31.6 & 39.8 \\
      UMT~\cite{deng2021unbiased} & 56.5 & 37.3 & 48.6 & 30.4 & 33.0 & 46.7 & 46.8 & 34.1 & 41.7 \\
      AT~\cite{li2022cross} & 60.0 & 49.0 & 63.6 & 38.8 & 45.0 & 53.9 & 45.1 & 33.9 & 49.0 \\
      CMT~\cite{cao2023contrastive} & 63.2  & 53.1 & 64.5 & 40.3 & 47.0 & 55.7 & 51.9 & \textbf{39.4} & 51.9 \\ 
     MILA~\cite{krishna2023mila} & 61.4  & 51.5 & 64.8 & 39.7 & 45.6 & 52.8 & \textbf{54.1} & 34.7 & 50.6 \\ 
      \midrule % <-- Midrule here 
      Ours & \textbf{64.8}  & \textbf{54.9} & \textbf{65.4} & \textbf{43.8} & \textbf{47.4} & \textbf{57.0} & 53.7 & 38.0 & \textbf{53.1}\\ 
      \midrule % <-- Midrule here
       Oracle (F-RCNN) & 50.3 & 40.7 & 61.3 & 32.5 & 43.1 & 49.8 & 35.1 & 28.6 & 42.7\\
      \bottomrule % <-- Bottomrule here
    \end{tabular*}
    }
    % }
  % \end{center}  
   \caption{ Domain adaptation from normal to adverse weather \textbf{(Cityscapes $\xrightarrow{}$ Foggy Cityscapes)}. The average precision (AP, $\%$) for all classes is reported. With VGG-16 as the backbone for fair comparison, our method achieves a new state-of-the-art result of \textbf{53.1\% mAP}, showing a gain of \textbf{+2.5} compared to MILA.}
    \label{tab:table2}
\end{table*}
%-------------------------------------------------------------------------

%-------------------------------------------------------------------------
% SIM10k
\begin{table}[t]
 \scriptsize
 \footnotesize	
  % \begin{center}
  \centering
  \addtolength{\tabcolsep}{-3.0pt}
    \begin{tabular}{l|c|r}
      \toprule % <-- Toprule here
      Method & Backbone & AP on Car\\
      \midrule % <-- Midrule here
      Source &    & 34.6 \\
      DA-Faster~\cite{chen2018domain} &  & 38.9 \\
      BDC-Faster~\cite{saito2019strong} &  & 31.8 \\
      SWADA~\cite{saito2019strong} &  & 40.1 \\
      MAF~\cite{he2019multi} &   & 41.1 \\
      SCDA~\cite{zhu2019adapting} &   & 43.0 \\
      CDN~\cite{li2020spatial} &  VGG-16 & 49.3 \\
      MeGA-CDA~\cite{vs2021mega} &   & 44.8 \\
      CADA~\cite{hsu2020every} &   & 49.0 \\
      UMT~\cite{deng2021unbiased} &   & 43.1\\
      D-adapt~\cite{jiang2021decoupled}&  & 50.3\\
      MILA& & 56.3\\
      Ours& & \textbf{57.0}\\
      \midrule % <-- Midrule here
      Oracle & & 69.7 \\
      \midrule % <-- Midrule here
      \midrule % <-- Midrule here

      Source &   & 41.8 \\
      CADA~\cite{hsu2020every} &  & 51.2 \\
      D-adapt~\cite{jiang2021decoupled} & ResNet-101 & 51.9\\
      MILA~\cite{krishna2023mila} &  & 58.2 \\
      Ours &  & \textbf{62.3} \\
       \midrule % <-- Midrule here
      Oracle &  & 70.4\\
      \bottomrule % <-- Bottomrule here
    \end{tabular}
  % \end{center}
  \caption{Domain adaptation from synthetic to real datasets \textbf{(Sim10k $\xrightarrow{}$ Cityscapes)}. Our method achieves the highest mAP of \textbf{62.3\%}, outperforming the best previous method, MILA, by \textbf{+4.1}.}

  %Following~\cite{saito2019strong}, We evaluate on the validation split of the Cityscapes and report the mAP on car.}
   \label{tab:table2_}
\end{table}
\begin{table}[t]
 %\scriptsize
 \footnotesize	
  % \begin{center}
  \centering
  \addtolength{\tabcolsep}{-5.0pt}
    \begin{tabular}{l|cccccc|r}
      \toprule % <-- Toprule here
      Method & bicycle & bird & car & cat & dog & person & mAP\\
      \midrule % <-- Midrule here
      Source & 32.5 & 12.0 & 21.1 & 10.4 & 12.4 & 29.9 & 19.7\\
       \midrule 
      DA-Faster~\cite{chen2018domain} & 31.1 & 10.3 & 15.5 & 12.4 & 19.3 & 39.0 & 21.2 \\
      SWADA~\cite{saito2019strong} & 36.4 & 21.8 & 29.8 & 15.1 & 23.5 & 49.6 & 29.4 \\
      MCAR~\cite{zhao2020adaptive} & 49.7 & 20.5 & 37.4 & 20.6 & 24.5 & 53.6 & 33.5 \\
      D-Adapt~\cite{jiang2021decoupled} & 52.4 & 25.4 & 42.3 & \textbf{43.7} & 25.7 & 53.5 & 40.5 \\
       MILA~\cite{krishna2023mila} & 59.1 & \textbf{28.5} & \textbf{49.8} & 28.3 & 35.7 & \textbf{66.3} & \textbf{44.6}\\
      \midrule % <-- Midrule here
      Ours & \textbf{63.4} & 24.6 & 48.7 & 27.9 & \textbf{38.3} & 64.1 & 44.5\\
      \midrule % <-- Midrule here
      Oracle & 44.2 & 35.3 & 31.9 & 46.2 & 40.9 & 70.9 & 44.6\\
      
      \bottomrule % <-- Bottomrule here
    \end{tabular}
  % \end{center}
  \caption{Domain adaptation from a real to artistic scenario (\textbf{PASCAL VOC $\xrightarrow{}$ Comic2k}), evaluated using ResNet-101 as the backbone. The average precision (AP, in $\%$) across all classes is reported.}
   \label{tab:table1_}
\end{table}
%-------------------------------------------------

\subsection{Performance Comparison}
We compare our proposed method with recently
published state-of-the-art methods, including SCL~\cite{shen2019scl}, SWADA~\cite{saito2019strong}, DM~\cite{kim2019diversify}, CRDA~\cite{xu2020exploring}, HTCN~\cite{chen2020harmonizing}, DA-Faster~\cite{chen2018domain}, MCAR~\cite{zhao2020adaptive}, D-Adapt~\cite{jiang2021decoupled}, MAF~\cite{he2019multi}, SCDA~\cite{zhu2019adapting}, CDN~\cite{li2020spatial}, MeGA-CDA~\cite{vs2021mega}, CADA~\cite{hsu2020every}, BDC-Faster~\cite{saito2019strong}, UMT~\cite{deng2021unbiased}, CMT~\cite{cao2023contrastive}, MILA~\cite{krishna2023mila}, and Adaptive Teacher (AT)~\cite{li2022cross}. %CMT~\cite{cao2023contrastive},
%Our implementation, as described in Sec.~\ref{sec:implementation_details}, builds upon the official code of \cite{li2022cross}. 
%
To ensure fair comparisons, we used the best reproducible results of AT under identical conditions. In our results, `Source' refers to the baseline model trained only on the source data without domain adaptation, while `Oracle' is trained and tested on the target domain. %For each method, we report mean Average Precision (mAP) and the improvement relative to Oracle.

\myparagraph{Adverse Weather Adaptation} 
The results of this setting are presented in Table~\ref{tab:table2}. Our method achieves the highest mAP in most categories. 
Notably, our method shows the largest improvement of $+8.6\%$ in the `train' class compared to AT~\cite{li2022cross}. This class has the fewest training samples, with only 504 instances.  This result indicates that the proposed memory module is particularly beneficial for classes with fewer training examples. We attribute this to the difficulty of aligning less populated classes, as it can be challenging to find suitable alignment targets. By storing all potential alignment targets in memory, our method effectively addresses this issue.

\myparagraph{Synthetic to Real Adaptation} 
Table~\ref{tab:table2_} presents the results for the car category in the Cityscapes dataset. Our method achieves a remarkable mAP of 62.3\%, outperforming the recent competitor D-Adapt~\cite{jiang2021decoupled} by a significant margin of 10.4\%. Moreover, it shows a 4.1\% improvement over our previous model, MILA~\cite{krishna2023mila}. This gain highlights the effectiveness of our proposed extensions to the MILA architecture—particularly the introduction of background feature alignment which have further reduced the domain gap and enabled this new state-of-the-art performance. In Sec.~\ref{sec:ablation2}, we analyze the effectiveness this extension individually to assess its contributions in detail.

\myparagraph{Real to Artistic Adaptation}
%Table~\ref{tab:table1_} shows the results of our real-to-artistic adaptation on Comic2k, while Table~\ref{tab:table3} and Table~\ref{tab:table1} present the results on Watercolor2k and Clipart1k, respectively. On Comic2k, our model achieves a notable mAP of 45.3\%, outperforming the recent competitor D-Adapt by 4.6\%.
Table~\ref{tab:table1_} shows the results of our real-to-artistic adaptation on Comic2k%, while Table~\ref{tab:table1} present the results on Clipart1k. 
Our model achieves a notable mAP of 44.5\%, outperforming the recent competitor D-Adapt by 4.0\%. 
%
%Similarly, our method achieves state-of-the-art results on Clipart1k, improving mAP by 2.2 compared to best score AT. 
%
%These results highlight the robustness of our approach across various artistic domains, demonstrating its effectiveness in bridging the domain gap consistently across different styles.
% 
These results consistently validate the effectiveness of aligning the most similar instances across domains in reducing the domain gap between different scenarios.

\begin{figure}[t!]
  \centering
  \includegraphics[width=\columnwidth]{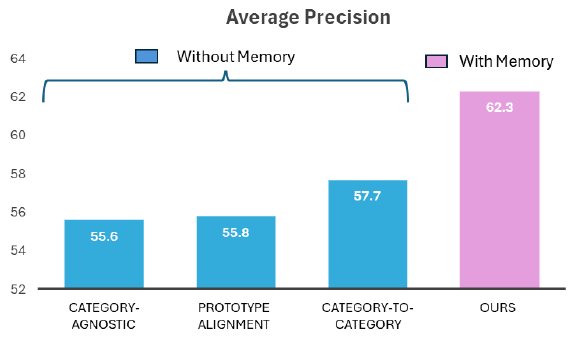}
  \caption{Impact of the proposed memory-based alignment module on detection accuracy is demonstrated. By enabling visually similar matches across batches, the memory-based approach enhances domain alignment, achieving a \textbf{4.6\%} performance improvement over the best-performing non-memory-based C2C method.}
    \label{fig:memory_module}
\end{figure}

\section{Analysis and Discussion}\label{sec:ablation1}
In this section, we provide a detailed analysis of our approach to assess the effectiveness of key components, examine parameter sensitivity, and visualize results. All analyses are conducted on the Sim10k$\xrightarrow{}$Cityscapes task.

\subsection{Ablation Study}\label{sec:ablation2}

\myparagraph{Effectiveness of memory module}
We evaluate the impact of the memory module by comparing our model's performance with and without it. When the memory module is absent, alignment relies solely on source instances within each mini-batch, and we assess three alignment strategies under this condition: category-agnostic, category-to-category, and prototype-based alignment.
% IJCV 0 -- Remove category-agnostic as not possible in Sim10K 
In the category-agnostic approach, a target instance (e.g., a red car) may align with any source instance, even if it belongs to a different category, such as 'Person.' C2C alignment ensures that a target instance (e.g., a red car) is matched exclusively with source instances of the same category (e.g., cars) within the mini-batch, if available. Prototype-based alignment matches the red car with a learned prototype for the 'car' category.
With the memory module, however, instance-level alignment improves significantly as it enables a red car in the target domain to match with a visually similar car (e.g., red) stored in memory, enhancing alignment quality. 
As shown in Fig.~\ref{fig:memory_module}, our memory-based method surpasses the non-memory-based C2C approach by 4.6\%, highlighting the effectiveness of memory-based alignment in reducing the domain gap and enhancing cross-domain performance.

% \myparagraph{Effect of different alignment losses}
% Here, we demonstrate the effectiveness of our proposed instance alignment loss, SimSupCon, and compare its performance with widely used Contrastive Loss (Con)\cite{tian2020contrastive} and Supervised Contrastive Loss (SupCon)\cite{khosla2020supervised}.
% %
% As illustrated in Fig.~\ref{fig:memory_module}, the detection performance reaches its highest point when SimSupCon is applied. This improvement shows that SimSupCon enhances domain alignment by placing greater emphasis on visually similar pairs, which is more effective than treating all pairs equally, as done in SupCon.

\begin{figure}[t!]
  \centering
  \includegraphics[width=\columnwidth]{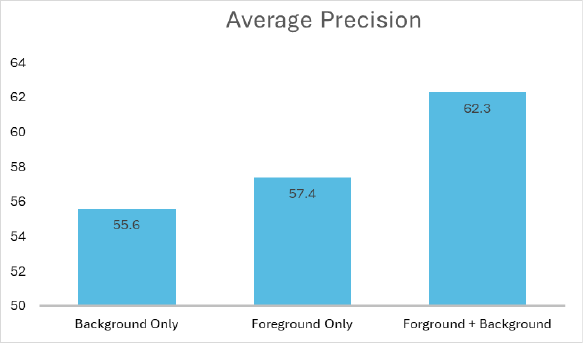}
  \caption{Performance comparison of alignment strategies: foreground-only, background-only, and combined alignment. The results demonstrate that aligning both foreground and background yields the best performance.}
    \label{fig:fig4}
\end{figure}

\myparagraph{Effect of foreground and background alignment}
In this part, we analyze the impact of the background alignment scheme we introduced as an extension of MILA. As shown in Fig.~\ref{fig:fig4}, we assess the model's performance across three settings: foreground-only alignment (original MILA, with $\lambda_3$=0), background-only alignment ($\lambda_2$=0), and both foreground and background alignment ($\lambda_2$ and $\lambda_3$ non-zero). The results indicate that aligning both the foreground and background leads to better performance than aligning only one of them, highlighting the benefits of combined alignment for improved domain adaptation.

\begin{figure}[t!]
  \centering
  \includegraphics[width=\columnwidth]{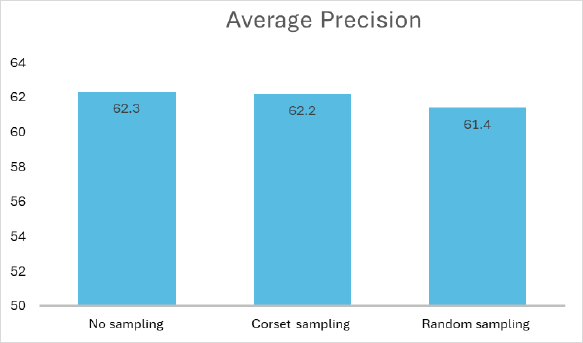}
  \caption{Comparison of subsampling methods for memory banks (\(\mathcal{M}_\text{bg}\) and \(\mathcal{M}_\text{fg}\)): greedy coreset selection versus random subsampling.}
    \label{fig:fig5}
\end{figure}

\myparagraph{Importance of memory subsampling} 
To optimize GPU memory usage and eliminate redundancy, we subsample the foreground and background memory banks during training. Figure~\ref{fig:fig5} compares two subsampling methods: greedy coreset selection and random subsampling. The results indicate that reducing the memory bank size by 50\% for the foreground and 70\% for the background using coreset selection achieves performance comparable to the original full-size memory banks (\(\mathcal{M}_\text{bg}\) and \(\mathcal{M}_\text{fg}\)), while outperforming random subsampling. By employing the subsampling strategy, we significantly reduce GPU memory overhead during training without compromising accuracy, ensuring efficient and effective domain alignment.

%To further analyze the importance of MILA's instance-level alignment module, we varied $\lambda_3$, which determines the weight for instance-level alignment in the training process and investigated its influence on the model’s performance. A large value means we are giving more importance to instance-level adaptation compared to image-level adaptation. If $\lambda_3$ is 0, the model performs image-level adaptation only. Note that this variant is different from what is presented in Fig.~\ref{fig:memory_module} under the name of `without our memory module' as it uses instance-level alignment with mini-batch samples. As shown in Table~\ref{tab:table22}, a high or low value of $\lambda_3$ can degrade the performance, and the peak accuracy is achieved when $\lambda_3$ is 0.1. Recalling that the default value of $\lambda_2$ in our experiment is 0.1, this result means that it is good to penalize the misalignment in image level and instance level to the same extent. Note that MILA achieves better mAP than the best performing existing method whose mAP is 59.7 (AT in Table~\ref{tab:table3}) even with slightly different values of $\lambda_3$.

\begin{figure*}[t!]
  \centering
  \includegraphics[width=\textwidth]{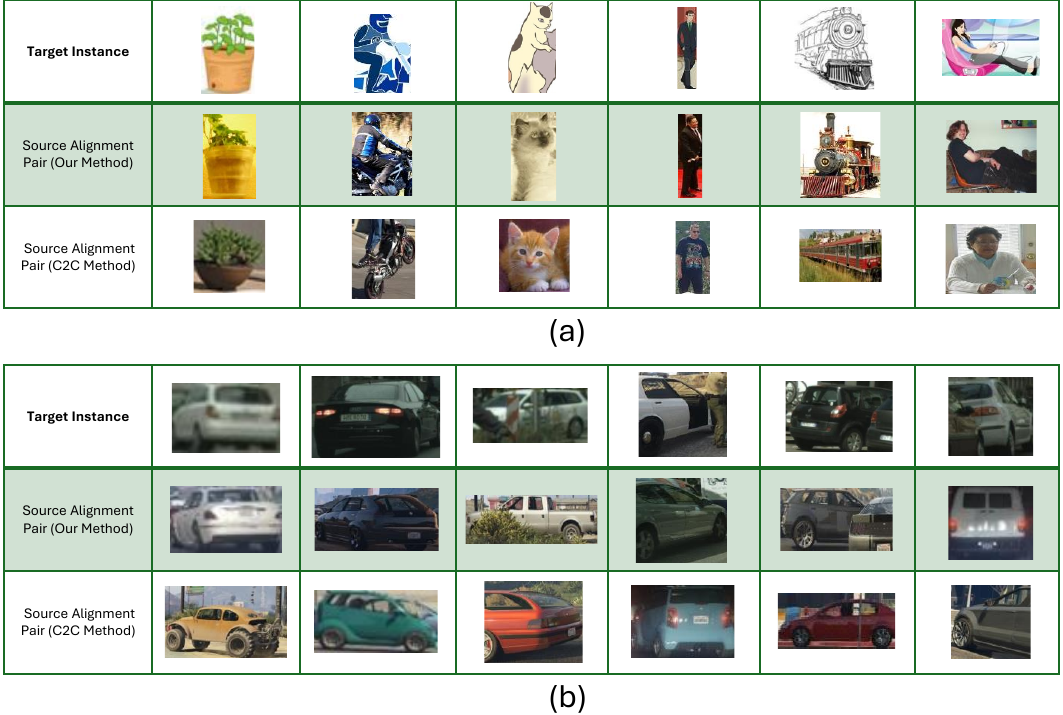}
  \caption{Visualization of instance pairs (a) Pascal VOC$\xrightarrow{}$Clipart1k (b) Sim10k$\xrightarrow{}$Cityscapes}
  \label{fig:fig6}
\end{figure*}

\subsection{Sensitivity Analysis}\label{sec:ablation}

\myparagraph{Sensitivity to confidence threshold $\delta$}
We varied the confidence threshold $\delta$ for filtering the noisy target bounding boxes and report the detection accuracy in Table~\ref{tab:table44}. The highest accuracy is obtained at the confidence threshold value of $0.8$. The result indicates that a very low value of $\delta$ allowed several noisy annotations to get aligned with source instances, and as a result, the mAP dropped. Similarly, a very high value of $\delta$ filters most of the generated instance, which makes the alignment less effective.

%----------------------------------------------
%Sensitivity to confidence threshold
\begin{table}[t]
 %\scriptsize
 \footnotesize	
  \centering
  %\addtolength{\tabcolsep}{-3.0pt}
    \begin{tabular*}{0.45 \textwidth}{@{\extracolsep{\fill}\quad}l|cccccc}
      \toprule % <-- Toprule here
      $\delta$ & 0.0 & 0.4 & 0.6 & 0.8 & 0.9   \\
      \midrule % <-- Midrule here
      mAP & 56.4 & 61.7 & 62.2 & \textbf{62.3}  & 60.9 \\
      \bottomrule % <-- Bottomrule here
    \end{tabular*}
  % \end{center}
   \caption{
   Effect of $\delta$, which controls the filtering of noisy predictions of target instances.}
   %Lower $\delta$ allows more noisy predictions for alignment, while too high $\delta$ allows only limited number of instances to be used for the alignment.}
    \label{tab:table44}
\end{table}
%----------------------------------------------

\myparagraph{Sensitivity to loss weights \( \lambda_2, \lambda_3 \) }
In this experiment, we test how sensitive our approach is to the values of \(\lambda_2\) and \(\lambda_3\), which balance foreground and background alignment losses. Tables~\ref{tab:table22_} and~\ref{tab:table22} show the model's performance with different values of \(\lambda_2\) and \(\lambda_3\), while keeping the other parameter fixed each time. Results show that very high or very low values for \(\lambda_2\) or \(\lambda_3\) reduce performance, and the best accuracy occurs when both \(\lambda_2\) and \(\lambda_3\) are set to 0.05.

%----------------------------------------------
%Sensitivity to loss weights\lambda_2
\begin{table}[t]
 %\scriptsize
 \footnotesize	
  % \begin{center}
  \centering
  %\addtolength{\tabcolsep}{-3.0pt}
    \begin{tabular*}{0.45\textwidth}{@{\extracolsep{\fill}\quad} l|ccccc}
      \toprule % <-- Toprule here
      $\lambda_2$ & 0.0 & 0.01 & 0.05 & 0.1    \\
      \midrule % <-- Midrule here
      mAP & 55.6 & 57.6 & \textbf{62.3} & 60.8  \\
      \bottomrule % <-- Bottomrule here
    \end{tabular*}
  % \end{center}
  \caption{
  % Sensitivity to instance-level alignment weight $\lambda_3$
  Effect of \( \mathcal{L}_{fg} \) on performance of our method. We vary $\lambda_2$ in [0.0,0.1] to control the impact of forground alignment (($\lambda_3=0.05$)). 
  }
    \label{tab:table22_}
\end{table}
%----------------------------------------------

%----------------------------------------------
%Sensitivity to loss weights\lambda_3
\begin{table}[t]
 %\scriptsize
 \footnotesize	
  % \begin{center}
  \centering
  %\addtolength{\tabcolsep}{-3.0pt}
    \begin{tabular*}{0.45\textwidth}{@{\extracolsep{\fill}\quad} l|ccccc}
      \toprule % <-- Toprule here
      $\lambda_3$ & 0.0 & 0.001 & 0.01 & 0.05 & 0.1   \\
      \midrule % <-- Midrule here
      mAP & 57.4 & 60.5 & 61.7 & \textbf{62.3} & 60.3  \\
      \bottomrule % <-- Bottomrule here
    \end{tabular*}
  % \end{center}
  \caption{
  % Sensitivity to instance-level alignment weight $\lambda_3$
  Effect of \( \mathcal{L}_{bg} \) on performance of our method. We vary $\lambda_3$ in [0,0.1] to control the impact of background alignment (($\lambda_2=0.05$)). 
  }
    \label{tab:table22}
\end{table}
%----------------------------------------------

\myparagraph{Sensitivity to number of retrieved pairs for alignment}
In this experiment, we determine the optimal number of source instances to retrieve from memory for alignment with each target instance (denoted by $K$), based on cosine similarity scores.
Table~\ref{tab:table33} shows that peak accuracy is achieved when only the most similar source instance is aligned with each target instance. This supports our claim that aligning the closest matching instances from the two domains allows our model to focus on adapting the domains effectively, without being affected by variations within the same category.

%----------------------------------------------
%Sensitivity to number of retrieved pairs
\begin{table}[t]
 %\scriptsize
 \footnotesize	
  % \begin{center}
  \centering
  %\addtolength{\tabcolsep}{-3.0pt}
    \begin{tabular*}{0.45 \textwidth}{@{\extracolsep{\fill}\quad} l|cccc}
      \toprule % <-- Toprule here
      % top-$K$ 
      $K$ & 1 & 10 & 30 & 100 \\
      \midrule % <-- Midrule here
      mAP & \textbf{62.3} & 56.9 & 57.4  & 56.5\\
      \bottomrule % <-- Bottomrule here
    \end{tabular*}
  % \end{center}
  \caption{Effect of varying $K$. Note that we retrieve top-$K$ similar source instance features from memory for a target instance.
  % Sensitivity to the number of retrieved memory pairs (top-$K$)
  }
    \label{tab:table33}
\end{table}

% \myparagraph{Effect of memory storage size $\gamma$}
% In the source domain dataset, there are hundreds of objects annotated in each category; storing all annotations would be memory inefficient. In this study, we vary the number of annotations to be stored in memory and report its effect on the detection accuracy.
% The parameter $\gamma$ denotes the ratio between number of instances to be stored in the memory and the total number of instances.
% The results are reported in Table~\ref{tab:table55}. 
% The result suggests that we can achieve competitive results even with significantly smaller memory sizes (recall that the mAP of the best performing existing method is 59.7 as shown in Table~\ref{tab:table3}). 
% This is very useful in real-world applications since it is not always viable to store all available instances in memory due to limited storage space and high computation cost.

% \begin{table}[t]
%  %\scriptsize
%  \footnotesize	
%   \centering
%   %\addtolength{\tabcolsep}{-3.0pt}
%     \begin{tabular*}{0.45 \textwidth}{@{\extracolsep{\fill}\quad}l|ccccccc}
%       \toprule % <-- Toprule here
%       $\gamma$ & 0.1 & 0.2 & 0.3 & 0.7 & 0.9 & 1.0   \\
%       \midrule % <-- Midrule here
%       mAP & 56.7 & 59.2 & 60.2 & 61.2 & 62.2 & \textbf{63.7}  \\
%       \bottomrule % <-- Bottomrule here
%     \end{tabular*}
%   % \end{center}
%  \caption{Effect of the memory size $\gamma$ on MILA. Lower $\gamma$ provides very limited number of candidates for retrieval, which results in retrieval of visually dissimilar instances. }
%     \label{tab:table55}
% \end{table}

\subsection{Visualization}\label{sec:visualization}

\myparagraph{Visualization of Alignment Pairs}  
Fig.~\ref{fig:fig6} (a) shows for predicted target instances (first row) how our method selects visually similar source pairs from memory that are well-suited for alignment.
For example, in the second example, our method aligns a target biker whose helmet and bike color match the source biker. In the fourth example, it aligns a target person wearing similar clothing to the source person, capturing fine visual details that are important for effective domain adaptation.
Fig.~\ref{fig:fig6} (b) presents results for car instances. Our method consistently selects cars with similar color and orientation to the target instances in first row, 1, 2, and 6 show back-facing cars, while examples 3, 4, and 5 display side-facing cars. In contrast, existing C2C method often select source instances that differ in color and orientation from the target.

These examples demonstrate our method’s advantage in identifying visually similar pairs for alignment over C2C. By aligning instances based on relevant visual similarities and ignoring unimportant differences, our model achieves more accurate cross-domain alignment.

\begin{figure*}[t!]
  \centering
  \includegraphics[width=\textwidth]{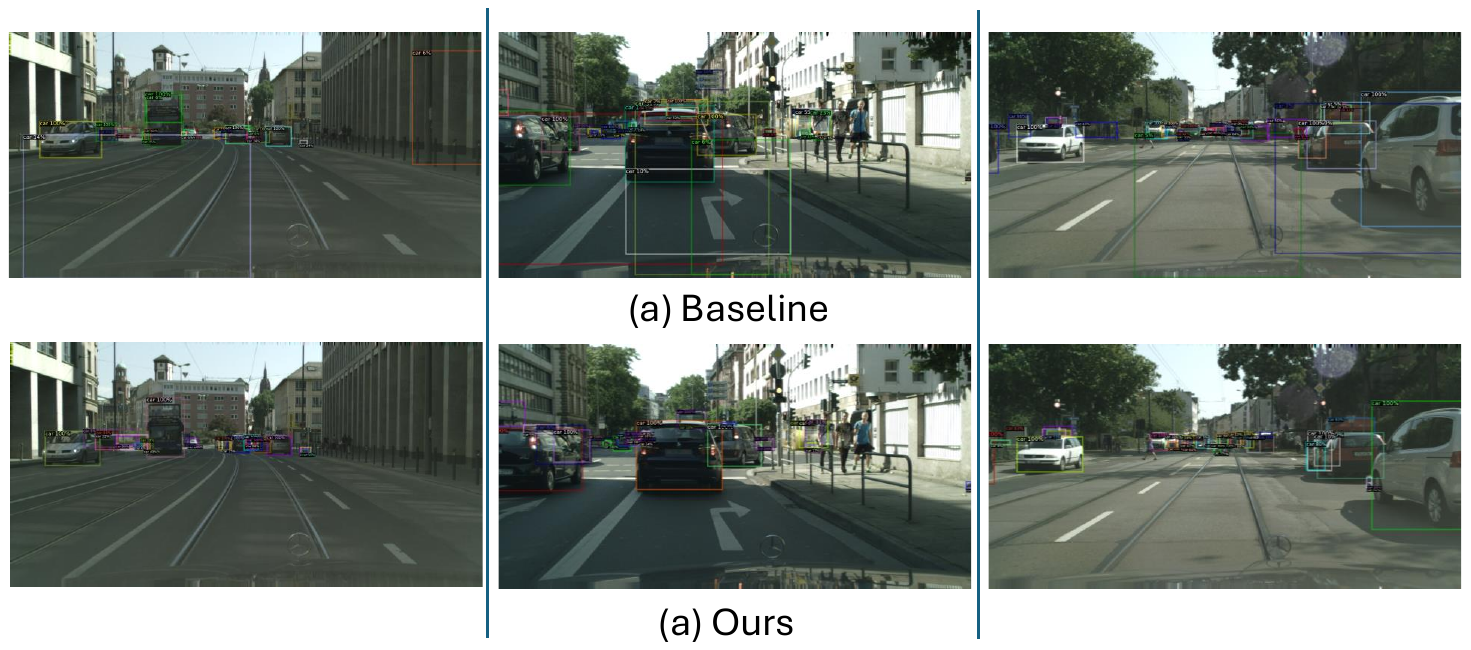}
  \caption{Visualization of detection results in the Synthetic-to-Real scenario, comparing our method with the previous state-of-the-art baseline, MILA~\cite{krishna2023mila}.}
  \label{fig:fig7}
\end{figure*}

\myparagraph{Qualitative detection results}  
Fig.~\ref{fig:fig7} presents examples of detection results for the Sim10k$\xrightarrow{}$Cityscapes task, comparing MILA with our method. The figure highlights that MILA struggles with accurate object localization and generates false positives. In contrast, the extension of MILA proposed in this work achieves more precise bounding box predictions, effectively reduces false positives, and accurately detects objects even in cases of severe occlusion.

\section{Conclusion}
In this paper, we experimentally validated that aligning visually similar pairs enhances domain alignment for cross-domain object detection, using a custom-built dataset. Building on this finding, we proposed a memory-based visually similar instance alignment framework for cross-domain object detection. Our framework stores features of foreground and background instances in separate memory modules, significantly larger than a mini-batch, enabling the selection of suitable source instances for alignment with target features across batches. This design enhances alignment by allowing the model to focus on domain-specific differences while minimizing irrelevant visual variations. 
Extensive experiments and analytical studies demonstrate the effectiveness of our approach, achieving state-of-the-art performance in cross-domain object detection.

\myparagraph{Acknowledgements}
Computational resource of AI Bridging Cloud
Infrastructure (ABCI) provided by National Institute of Advanced
Industrial Science and Technology (AIST) was used.

\section*{Declarations}
\myparagraph{Conflict of Interest}
The authors declare that they have no conflict of interest.

\myparagraph{Funding}
Not applicable.

\myparagraph{Data Availability}
The data and public code availability: \url{https://github.com/hitachi-rd-cv/MILA}

\myparagraph{Ethical Approval}
Not applicable.

\myparagraph{Consent to Participate}
Not applicable.

\myparagraph{Consent for Publication}
Not applicable.

\bibliography{sn-bibliography}% common bib file

\end{document}